\documentclass[a4paper, 10pt, conference]{ieeeconf}      
\usepackage{FG2021,amsmath,amssymb,epsfig, subcaption,multirow,array,color,hyperref,ctable}

\FGfinalcopy 

\IEEEoverridecommandlockouts                              
\def\FGPaperID{327} 

\title{\LARGE \bf
Unimodal Face Classification with Multimodal Training
}

\author{\parbox{16cm}{\centering
    {\large Wenbin Teng$^1$ and Chongyang Bai$^2$}\\
    {\normalsize
    $^1$ Boston University\\
    $^2$ Dartmouth College}
    }
}

\begin{document}

%
%
%




\IEEEoverridecommandlockouts\pubid{\makebox[\columnwidth]{978-1-6654-3176-7/21/\$31.00~\copyright{}2021 IEEE \hfill}
\hspace{\columnsep}\makebox[\columnwidth]{ }}

\ifFGfinal
\thispagestyle{empty}
\pagestyle{empty}
\else
\author{Anonymous FG2021 submission\\ Paper ID \FGPaperID \\}
\pagestyle{plain}
\fi
\maketitle

\newcommand{\edit}[1]{\textcolor{red}{#1}}

\begin{abstract}
Face recognition is a crucial task in various multimedia applications such as security check, credential access and motion sensing games. However, the task is challenging when an input face is noisy (e.g. poor-condition RGB image) or lacks certain information (e.g. 3D face without color). In this work, we propose a \underline{M}ultimodal \underline{T}raining \underline{U}nimodal \underline{T}est (MTUT) framework for robust face classification, which exploits the cross-modality relationship during training and applies it as a complementary of the imperfect single modality input during testing. Technically, during training, the framework (1) builds both intra-modality and cross-modality autoencoders with the aid of facial attributes to learn latent embeddings as multimodal descriptors, (2) proposes a novel multimodal embedding divergence loss to align the heterogeneous features from different modalities, which also adaptively avoids the useless modality (if any) from confusing the model. This way, the learned autoencoders can generate robust embeddings in single-modality face classification on test stage. We evaluate our framework in two face classification datasets and two kinds of testing input: (1) poor-condition image and (2) point cloud or 3D face mesh, when both 2D and 3D modalities are available for training. We experimentally show that our MTUT framework consistently outperforms ten baselines on 2D and 3D settings of both datasets\footnote{Code can be found in the following url: \url{https://github.com/wbteng9526/mtut_fr}}.

\end{abstract}

\section{INTRODUCTION}

\label{sec:intro}

Recent face recognition technologies have achieved good performance with the advances of deep learning. Convolutional Neural Networks (CNNs) have become the state-of-the-art in current 2D image market \cite{taigman2014deepface, sun2014deep,schroff2015facenet,simonyan2014very}. However, the deep CNNs trained from 2D face images still suffer from effects of large variations in image properties (e.g. illumination and sharpness) and face properties (e.g. pose and expression) \cite{xu2017evaluation}. 
In addition, recognizing human faces based on single 2D information would potentially increase the security risk as the system would be fooled by images lack of voxel information \cite{yi2014face}.
Compared to 2D images, 3D data (e.g. point cloud and mesh) would overcome the problem as it is intrinsically invariant to pose, illumination, and sharpness \cite{xu2017evaluation}. Moreover, it provides a detailed geometric description of the muscle, shape, contour, and key points of faces.
Compared to 2D RGB images, 3D point clouds contain rarer texture and color information of faces. However, point clouds are usually noisy, sparse, unorganized \cite{xu2004automatic} and not as easy to acquire as 2D images are, 
making modeling difficult to setup.

\begin{figure}[htb!]
\centering
\begin{subfigure}{.5\textwidth}
  \centering
  \includegraphics[width=\linewidth]{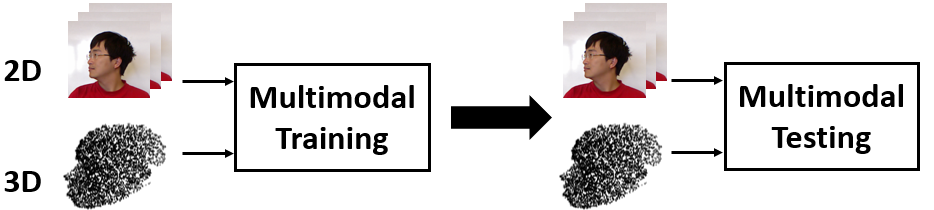}
  \vspace{-5mm}
  \label{fig:sub1}
  \caption{Multimodal training, \emph{multimodal} testing}
\end{subfigure}
\hfill
\vspace{5mm}
\begin{subfigure}{.5\textwidth}
  \centering
  \includegraphics[width=\linewidth]{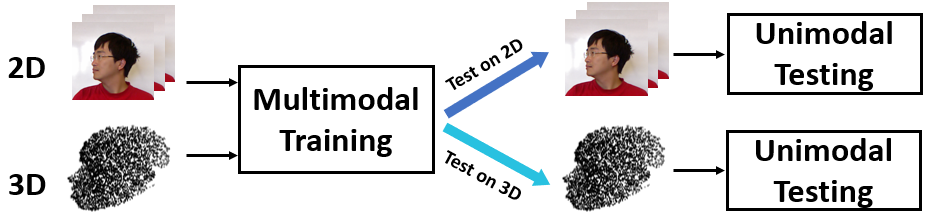}
  \vspace{-5mm}
  \label{fig:sub2}
  \caption{Multimodal training, \emph{unimodal} testing}
\end{subfigure}
\caption{Different training and testing procedures for different systems. (a) depicts both training and testing with multiple modalities. (b) depicts training with multiple modalities but only testing with single modality} 
\vspace{-4mm}
\label{fig:multi_uni_diagram}
\end{figure}

Therefore, one next developing direction is the effective combination of both 2D and 3D modalities, tapping each other's strength to make up for individual weak points.
Indeed, there are some works involving both 3D and 2D modalities \cite{kakadiaris20173d,xu2017evaluation}. Those models are evaluated based on the availability of both modalities. In practice, however, the acquirement of high-quality 3D data can be costly \cite{wang2019dynamic}; on the other hand, 
the low-quality 2D images (e.g. light condition, camera shaking) can harm model performances. Therefore, not both modalities are available during testing.

\begin{figure*}[h]
\begin{center}
  \includegraphics[width=\textwidth]{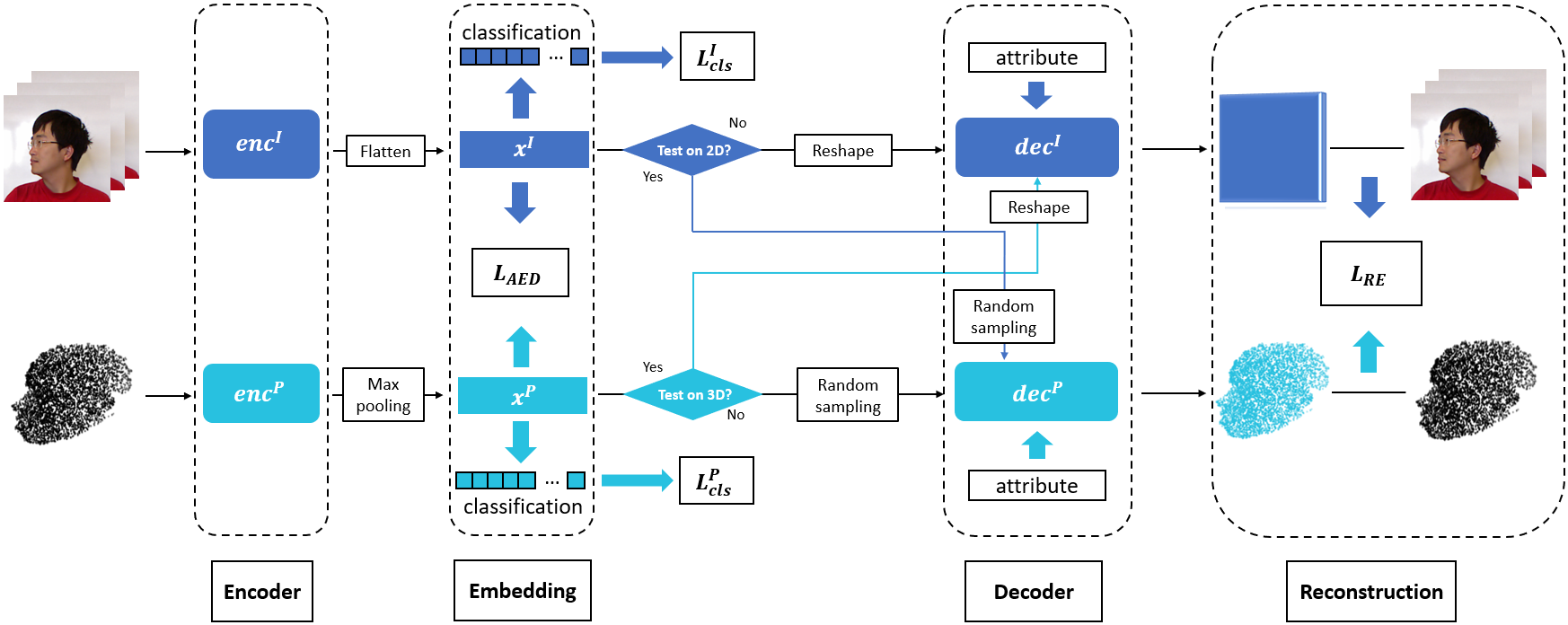}
\end{center}
\vspace{-2mm}
  \caption{An overview of proposed MTUT face recognition framework
  : both 2D and 3D inputs are encoded into individual modality embeddings. 
  Divergence between both embeddings ($\mathcal{L}_{AED}$) will be minimized adaptively through measuring the classification loss ($\mathcal{L}_{cls}^I$ or $\mathcal{L}_{cls}^P$) of each modality. Based on the type of testing modality, we use certain decoder to reconstruct 2D and 3D inputs from feature embeddings. For example, when only 2D input is available for test, both 2D and 3D embeddings will be reconstructed to 3D input during training. Reconstruction loss ($\mathcal{L}_{RE}$) is also optimized.} 
  \label{fig:autoencoder}
  \vspace{-5mm}
\end{figure*}

In this work, we focus on the \emph{multimodal training unimodal testing (MTUT)} problem. The difference between our MTUT and other multimodal learning frameworks is illustrated in Fig. \ref{fig:multi_uni_diagram}. We assume that both 2D and 3D modalities are available for training, while two scenarios could happen on test stage: (1) only 2D images are available 
, and (2) only 3D point clouds exist. 
In practice, if we have no prior knowledge of the available modality during testing, training models for both scenarios is recommended. We propose our MTUT framework which trains both \emph{cross-modal} and \emph{intra-modal} autoencoders. Assume $\mathcal{M}$ and $\mathcal{A}$ are the \emph{missing} and \emph{available} modalities during test, respectively.
During training, both modalities are encoded into descriptive feature embeddings, which are then decoded to reconstruct the missing modality $\mathcal{M}$. 
To improve the reconstruction, the decoding process is guided by the face attributes of each individual (e.g. bangs, black hair, and goatee). We apply a pre-trained face attribute classification model on CelebA dataset \cite{liu2015faceattributes} to obtain 40 attributes and combine with extracted feature embeddings.
The \emph{cross-modal autoencoder} enables the encoded embedding of available test modality $\mathcal{A}$ to effectively carry the information of the missing test modality $\mathcal{M}$. 
The \emph{intra-modal autoencoder} aims to give the latent embedding of $\mathcal{A}$ more descriptive and reliable information. The latent embedding of $\mathcal{A}$ is also optimized for the face classification task.

Since modalities of a face should be intrinsically aligned, we aim to minimize the divergence between the latent embeddings of both modalities. However, sometimes a noisy modality may play a negative role in classification (\cite{hori2017attention,bai2020m2p2}); therefore, in order to avoid the interference of the noisy latent embedding, we propose to optimize an \emph{adaptive embedding divergence (AED) loss}. The idea is to evaluate the training classification performance of both embeddings to adaptively decide whether to perform feature divergence optimization. 
Therefore, when training, our objectives consist of single modality classification loss, AED loss with the avoidance of negative transfer and the reconstruction loss of autoencoders.  


In sum, we make the following contributions: (1) We propose a novel framework achieving multimodal training and unimodal testing (MTUT) in face recognition. Experiments show that the framework is superior to SOTA systems on two datasets in both 2D and 3D settings; (2) We establish face attribute guided intra-model and cross-modal autoencoders to learn embeddings containing information of both the available and missing modalities during test; (3) We develop an adaptive embedding divergence (AED) loss to generate the multimodal embedding while avoiding interference from any potential noisy modality.

\section{Proposed Method}

\subsection{Architecture} \label{sec:3.1}

Fig. \ref{fig:autoencoder} shows our proposed multimodal training unimodal testing (MTUT) architecture, which consists of intra-modal and cross-modal autoencoders trained by three kinds of losses: modality classification loss $\mathcal{L}_{cls}^{\mathcal{M}} (\mathcal{M} \in \{I,P\})$, adaptive embedding divergence loss $\mathcal{L}_{AED}$ and reconstruction loss $\mathcal{L}_{RE}$. The encoders are shown on the left part of Fig. \ref{fig:autoencoder}: we denote 2D RGB image input as $\mathbf{I} \in \mathbb{R}^{H\times W \times C}$. A 2D encoder is used to extract the 2D face embedding: $\mathbf{x}^I \in \mathbb{R}^q$ from the mean pooling layer connected to the last convolution layer. 
For the corresponding 3D point cloud input\footnote{We treat mesh as point cloud, too.} $\mathbf{P} \in \mathbb{R}^{n \times F}$ with $n$ points each represented by an $F$-dimensional feature vector, we employ PointNet \cite{qi2017pointnet} architecture, which extracts 3D point features through graph convolutional  network (GCN) and aggregate to feature embedding $\mathbf{x}^P \in \mathbb{R}^q$ through all points by max-pooling. We denote 2D and 3D encoders as $enc^I$ and $enc^P$ so that:

\begin{equation}
    \mathbf{x}^I = enc^I(\mathbf{I}),\; \mathbf{x}^P = enc^P(\mathbf{P})
\end{equation}

Once the whole model is trained, either embedding $\mathbf{x}^P$ or $\mathbf{x}^I$ is used for face classification during test stage, depending on which modality is available. 
During training, only the modality available for test is optimized by the following cross entropy loss for face recognition:

\begin{equation} \label{eq:class}
\begin{split}
    \mathcal{L}^I_{cls} &= -\mathbb{E}_{\mathbf{x}^I}\log\mathbb{P}(\mathbf{y}|\mathbf{x}^I), \\
    \text{or} \\
    \mathcal{L}^P_{cls} &= -\mathbb{E}_{\mathbf{x}^P}\log\mathbb{P}(\mathbf{y}|\mathbf{x}^P),
\end{split}
\end{equation}

\noindent where $\mathbb{E}_{\mathbf{x}}$ represents the expected value operator with respect to the distribution of $\mathbf{x}$,  $\mathbb{P}(\mathbf{y}|\mathbf{x}^P)$ represents conditional probability operator and $\mathbf{y}$ stands for the identities of faces.

However, single modality classification would sometimes be unreliable since the encoder network may generate embedding that not in perfect alignment. To tackle this problem, 
we propose an adaptive embedding divergence (AED) loss that ornaments feature divergence with a regularization parameter $\rho$ to measure the scale that feature divergence is going to be optimized. Noted as $\mathcal{L}_{AED}$, the loss is defined as: 

\begin{equation} \label{eq:div}
    \mathcal{L}_{AED} = \rho \left\|\mathbf{x}^I - \mathbf{x}^P\right\|_{2}
\end{equation}

\noindent$\left\| \cdot \right\|_{2}$ represents the L2-norm. Details of the regularization parameter $\rho$ will be discussed in Section \ref{sec:3.4}. 

The right part of Fig. \ref{fig:autoencoder} illustrates the decoder network. To empower reconstruction, we combine encoded feature embedding with face attributes extracted through \cite{he2018harnessing}. 
We combine feature embedding with the probability of an individual face being classified as certain attributes. 
Suppose the latent feature after attribute combination is denoted as $\mathbf{x}^{\prime I}$ and $\mathbf{x}^{\prime P}$, then:

\begin{equation} \label{eq:attr}
\begin{split}
    \mathbf{x}^{\prime I} = f(\mathbf{x}^I,\mathbf{a}),\;
    \mathbf{x}^{\prime P} = f(\mathbf{x}^P,\mathbf{a}),
\end{split}
\end{equation}

\noindent where $\mathbf{a} \in \mathbb{R}^a$ is the attribute vector specifically belongs to each individual face and $f$ is a fully-connected layer for dimension reduction, which takes the concatenation of attribute vector and modality embedding to remove the redundant information lied in input space. Based on the type of modality available on testing stage, we propose different reconstruction processes during training. 


\noindent{\textbf{Case 1: 3D modality is available and 2D modality is missing for testing.}} During training, both 2D and 3D feature embeddings $\mathbf{x}^{\prime I}$ and $\mathbf{x}^{\prime P}$ will be the input of decoder network to generate $\hat{\mathbf{I}}$ and $\hat{\mathbf{P}}$ respectively as the reconstruction of $\mathbf{I}$:

\begin{equation} \label{eq:deci}
    \begin{split}
        \hat{\mathbf{I}} = dec^I(\mathbf{x}^{\prime I}),\;
        \hat{\mathbf{P}} = dec^I(\mathbf{x}^{\prime P})
    \end{split}
\end{equation}

\noindent where $dec^I$ denotes the decoder network. The reconstruction loss is defined as:

\begin{equation} \label{eq:relossP}
    \mathcal{L}_{RE}^I=\left\|\hat{\mathbf{I}}-\mathbf{I}\right\|_{2}+\left\|\hat{\mathbf{P}}-\mathbf{I}\right\|_{2}
\end{equation}

\noindent{\textbf{Case 2: 2D modality is available and 3D modality is missing for testing.}} During training, the decoder network generates $\hat{\mathbf{I}}$ and  $\hat{\mathbf{P}}$ as reconstruction $\mathbf{I}$. Therefore:

\begin{equation} \label{eq:decp}
    \begin{split}
        \hat{\mathbf{I}} = dec^P(\mathbf{x}^{\prime I}),\;
        \hat{\mathbf{P}} = dec^P(\mathbf{x}^{\prime P})
    \end{split}
\end{equation}

\noindent where $dec^P$ is the decoder network. Then the reconstruction loss is defined as:

\begin{equation} \label{eq:relossI}
    \mathcal{L}_{RE}^P=\left\|\hat{\mathbf{I}}-\mathbf{P}\right\|_{2}+\left\|\hat{\mathbf{P}}-\mathbf{P}\right\|_{2}
\end{equation}


The full objective function of our proposed architecture will be discussed in Section \ref{sec:3.5}. 

\subsection{Adaptive Embedding Divergence (AED) Loss} \label{sec:3.4}
The encoded embeddings from both 2D image and 3D point cloud should provide complementary descriptions of face input from different perspectives.
However, 2D features sometimes fail to include comprehensive geometric representations, whereas 3D features, containing most of the stereoscopic information, are sometimes unable to provide 2D appearance and attributes such as the color of skin and hair.
It is important to note that sometimes the shortcomings of one modality would outperform the other (e.g. noisy 3D point cloud, low-quality 2D image), which would weaken the role of the modality played during multimodal training. 

Therefore, it is necessary to measure and compare the quality of embeddings from both modalities in the classification network and then decide whether to optimize the embedding divergence between the two.
For example, if we are going to test 3D input, we would measure $\Delta \mathcal{L} = \mathcal{L}^P_{cls} - \mathcal{L}^I_{cls}$, the difference between the classification loss of 3D and 2D modality embeddings. 
A positive $\Delta \mathcal{L}$ means a better performance of 2D modality embedding in the classification network, illustrating a higher quality of image input. Thus, we are more inclined to bring extracted 3D modality embedding closer to the 2D one. Larger $\Delta \mathcal{L}$ means that 2D modality has much better performance than 3D, thus we expect 3D feature to mimic 2D more strictly.
Assume $\mathcal{M}$ is the missing modality during test, 
we are inspired by \cite{abavisani2019improving} to define an adaptive regularization parameter $\rho^{\mathcal{M}}$ to be:

\begin{equation}
    \rho^{\mathcal{M}} =\left\{\begin{array}{cl}
e^{\beta \Delta_{\mathcal{M}} \mathcal{L}} - 1, & \Delta_{\mathcal{M}} \mathcal{L} > 0 \\
0, & \text{otherwise}
\end{array}\right.
\end{equation}

\noindent where $\beta > 0$ is a  hyper-parameter controlling the impact from the loss difference. Particularly, if $\mathcal{M} = I$ (missing image for test), then $\Delta_{I} \mathcal{L} = \mathcal{L}^{I}_{cls} - \mathcal{L}^P_{cls}$, whereas if $\mathcal{M} = P$ (missing 3D modality for test), then $\Delta_{P} \mathcal{L} = \mathcal{L}^{P}_{cls} - \mathcal{L}^I_{cls}$.

\subsection{Full Objective of MTUT Network} \label{sec:3.5}

Our objective function considers a weighted sum of the classification loss ($\mathcal{L}_{cls}^I$ or $\mathcal{L}_{cls}^P$), reconstruction loss from both 2D and 3D autoencoders ($\mathcal{L}_{RE}^I$ or $\mathcal{L}_{RE}^P$), and the adaptive embeddinng divergence loss ($\mathcal{L}_{AED}$):




\begin{equation} \label{eq:obj}
    \mathcal{L} =\left\{\begin{array}{cl}
\mathcal{L}_{cls}^P + \lambda_1 \mathcal{L}_{RE}^I + \lambda_2 \mathcal{L}_{AED}, & \mathcal{M}=I \\

\mathcal{L}_{cls}^I + \lambda_1 \mathcal{L}_{RE}^P + \lambda_2 \mathcal{L}_{AED}, & \mathcal{M}=P
\end{array}\right.
\end{equation}

\noindent $\lambda_1$, $\lambda_2$ are weights balancing the three loss components. We adjust the value of $\lambda_1$,$\lambda_2$ by grid search.

Overall, our proposed networks are trained to improve the representations of latent feature embedding and further boost classification accuracy during training stage. During testing mode, each modality will be evaluated separately.

\begin{table*}[!t]
    \centering
    \begin{tabular}{|c|c|c|c|c|c|}
        \hline
        \multirow{2}{*}{Test Modality} & \multirow{2}{*}{Method} & \multicolumn{2}{c|}{Kinect} & \multicolumn{2}{c|}{CASIA}\\
        \cline{3-6}
        & & Accuracy (\%) & F1-score (\%) & Accuracy (\%) & F1-score (\%)\\
        \hline

        \multirow{10}{*}{2D} &
        VGG-11 \cite{simonyan2014very} & 81.04 & 80.84 & 90.87 & 91.05\\
        &\textbf{VGG-11 \cite{simonyan2014very}+MTUT} & \textbf{84.29} & \textbf{82.92} & \textbf{92.00} & \textbf{92.28}\\
        \cline{2-6}
        &ResNet-18 \cite{he2016deep} & 94.87 & 94.65 & 94.70 & 94.55\\
        &\textbf{ResNet-18 \cite{he2016deep}+MTUT} &\textbf{97.43} & \textbf{97.17} & \textbf{95.24} & \textbf{94.87}\\
        \cline{2-6}
        &FaceNet \cite{schroff2015facenet} & 90.66 & 90.56 & 91.57 & 91.43\\
        &\textbf{FaceNet \cite{schroff2015facenet}+MTUT} & \textbf{93.12} & \textbf{92.97} & \textbf{93.42} & \textbf{93.28}\\
        \cline{2-6}
        &DeepID \cite{sun2014deep} & 63.19 & 63.19 & 76.54 & 76.12\\
        &\textbf{DeepID \cite{sun2014deep}+MTUT} & \textbf{67.50} & \textbf{67.49} & \textbf{78.24} & \textbf{77.82}\\
        \cline{2-6}
        &DeepFace \cite{taigman2014deepface} & 75.00 & 74.55 & 74.38 & 73.49\\
        &\textbf{DeepFace \cite{taigman2014deepface}+MTUT} & \textbf{80.19} & \textbf{79.21} & \textbf{75.45} & \textbf{74.18}\\
        \cline{1-6}
        \multirow{2}{*}{3D} &
        PointNet \cite{qi2017pointnet} & 79.49 & 79.17 & 82.49 & 81.31\\
        &\textbf{PointNet \cite{qi2017pointnet}+MTUT} & \textbf{86.58} &\textbf{86.34} & \textbf{89.84} & \textbf{89.41}\\
        \hline
    \end{tabular}
    \vspace{1mm}
    \caption{Face recognition on Kinect \cite{min2014kinectfacedb} and CASIA \cite{biometricsidealtest}.}
    \label{tab:all_result}
    \vspace{-3mm}
\end{table*}

\section{Experiments}
\vspace{-1mm}
\subsection{Datasets} \label{sec:4.1}
\vspace{-1mm}
\noindent\textbf{KinectFaceDB} We evaluate our proposed model on \emph{KinectFaceDB} \cite{min2014kinectfacedb}. There are 52 individual faces. For each person, 2D RGB images and 3D point clouds will be used by our in method and the baselines. 
There are in total of 728 2D images and 728 3D point clouds.

\noindent\textbf{CASIA 3D Face Database} Portions of the research in this paper use the CASIA-3D FaceV1 \cite{biometricsidealtest} collected by the Chinese Academy of Sciences. It contains both 2D and 3D database of 123 people with 4,624 scans. 
The 3D dataset provides 3D mesh data, including point coordinates, color, and normal vectors. In correspondence with our proposed situation, we only use the 3D coordinates as our input.

\subsection{Pre-processing} \label{sec:4.2}
\begin{figure}[t]
  \includegraphics[width=\linewidth]{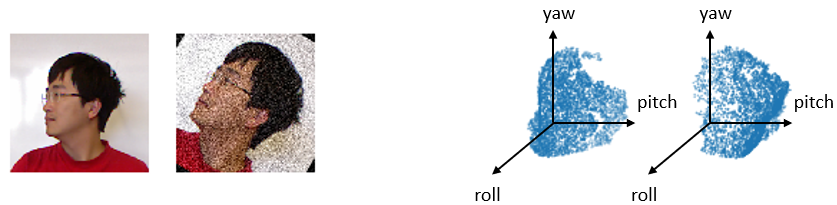}
  \vspace{-3mm}
  \caption{Examples of 2D and 3D data augmentation.}
  \label{fig:augment}
  \vspace{-5mm}
\end{figure}

There are a total of 728 human faces in Kinect database \cite{min2014kinectfacedb}, and we split it into 50\% for training, 20\% for validation and 30\% for testing. In CASIA 3D \cite{biometricsidealtest}, there are 4,624 scans and we split them into 60\%, 20\% and 20\% for train, validation and test, respectively. 2D image data is resized to 224$\times$224; 3D coordinates are extracted from original meshes and we randomly sample 4,000 points from more than 60,000 points. After that we augment 2D images through random cropping, adding Gaussian noise and rotation and augment 3D point clouds through rotation on roll, pitch and yaw axis. Fig. \ref{fig:augment} visualizes an example that compares before and after augmentation.

\subsection{Implementation Details}
\vspace{-1mm}
In design of our method, we adopt the architectures of ResNet-18 \cite{he2016deep} and PointNet \cite{qi2017pointnet} as the backbone network of our 2D and 3D modality. 
To optimize Eqn. (\ref{eq:obj}), weight $\lambda_1$ and $\lambda_2$ are set to be 0.001 and 0.1, respectively. We use Adam optimizer and betas is set to be (0.9, 0.999). We set the default learning rate of $10^{-3}$ with a 10$\times$ reduction when the loss is saturated. We set batch size of training to be 32 and whole network will be trained on both dataset for 30 epochs. The whole architecture is implemented in PyTorch. 

\subsection{Results}

\begin{table}[t]
    \centering
    \begin{tabular}{|l|c|c|c|}
    \hline
    Method & Test Modality & Kinect & CASIA \\
    \hline
    DCC-CAE \cite{dumpala2019audio} & 2D & 91.67 & 92.64\\
    SSA \cite{abavisani2019improving} & 2D & 93.59 & 95.03\\
    \textbf{MTUT (ours)} & 2D & \textbf{97.43} & \textbf{95.24} \\
    \hline
    DCC-CAE \cite{dumpala2019audio} & 3D & 73.72 & 82.81\\
    SSA \cite{abavisani2019improving} & 3D & 85.89 & 89.08\\
    \textbf{MTUT (ours)} & 3D & \textbf{85.90} & \textbf{89.84}\\
    
    \hline
    \end{tabular}
    \caption{Comparison with other state-of-the-art multimodal learnng methods on Kinect \cite{min2014kinectfacedb} and CASIA \cite{biometricsidealtest}. The backbone method for 2D and 3D modality are ResNet-18 \cite{he2016deep} and PointNet \cite{qi2017pointnet}. The scores are reported as accuracy.}
    \label{tab:multi}
    \vspace{-5mm}
\end{table}

    

This section shows the result of our proposed model on KinectFaceDB \cite{min2014kinectfacedb} and CASIA 3D Face Database \cite{biometricsidealtest}. We extend the state-of-the-art methods to incorporate the other face modality to illustrate that the performance of single modality would benefit from multimodal learning. Table \ref{tab:all_result} compares evaluation accuracy and F1-score in terms of unimodal learning vs. multimodal learning. Table \ref{tab:multi} compares our MTUT model with some other state-of-the-art learning models.

\noindent\textbf{Performance on KinectFaceDB.} Our proposed method shows great improvement in face recognition accuracy compared with current state-of-the-art architectures on KinectFaceDB \cite{min2014kinectfacedb}. Compared with backbone model, we boost the performance from 94.87\% (ResNet-18) to 97.43\% in 2D testing mode and boost performance from 82.59\% (PointNet) to 89.84\% in 3D testing mode. 

\noindent\textbf{Performance on CASIA 3D Face Database.} Our proposed method also improves accuracy on CASIA 3D Face Database \cite{biometricsidealtest}. In more details, MTUT boosts the accuracy of ResNet-18 by 0.54\% in 2D testing mode and PointNet by 7.35\%. Our MTUT model also outperforms other state-of-the-art multimodal learning methods.



\section{Conclusion}

We propose a new face recognition framework that achieves multimodal training and unimodal test (MTUT) when only one of the modalities is available. Our models learn intra-modal and inter-modal autoencoders for each modality with the aid of facial attributes and avoid negative information transfer through proposed adaptive embedding divergence loss. Experiments show that our method improves the performance of baselines remarkably. One future research direction is to 
generalize the framework towards other multimodal tasks (e.g. sentiment analysis and emotion prediction in videos).





{\small
\bibliographystyle{ieee}
\bibliography{egbib}
}

\end{document}


%
%
%




\IEEEoverridecommandlockouts\pubid{\makebox[\columnwidth]{978-1-6654-3176-7/21/\$31.00~\copyright{}2021 IEEE \hfill}
\hspace{\columnsep}\makebox[\columnwidth]{ }}

\ifFGfinal
\thispagestyle{empty}
\pagestyle{empty}
\else
\author{Anonymous FG2021 submission\\ Paper ID \FGPaperID \\}
\pagestyle{plain}
\fi
\maketitle

\newcommand{\edit}[1]{\textcolor{red}{#1}}

\begin{strip}
  \centering
  \includegraphics[width=\textwidth]{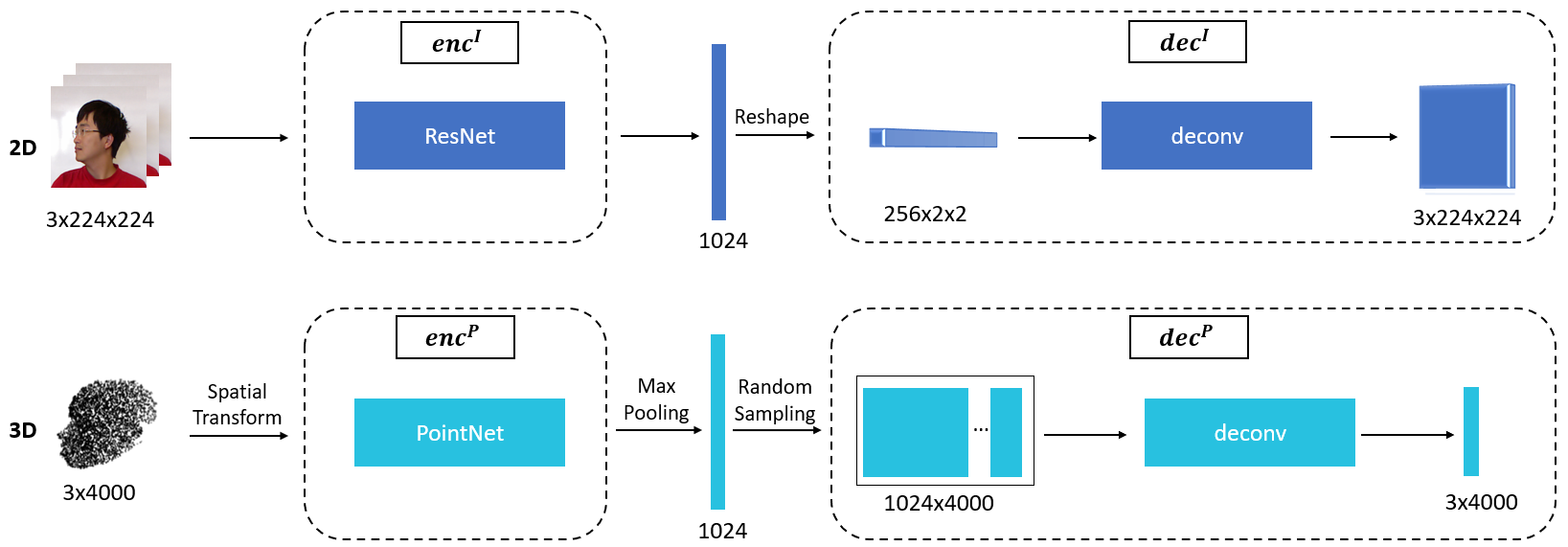}
  \captionof{figure}{An overview of 2D (top row) and 3D (bottom row) autoencoder networks: 2D encoder applies ResNet \cite{he2016deep} and 3D applies PointNet \cite{qi2017pointnet}. Both 2D and 3D decoder networks apply multiple deconvolution layers. The decoder networks are trained to reconstruct the missing modality in testing mode. For example, if 3D point cloud is missing during testing, 3D decoder is trained to reconstruct 3D point cloud.} 
  \label{fig:decoder}
\end{strip}

\noindent Our supplementary document is organized as follows. We introduce the details of our 2D and 3D autoencoders in Section \ref{sec:2D_autoencoder} and Section \ref{sec:3D_autoencoder}. Section \ref{sec:visual} visualizes some other evaluation results and ablation studies in addition to our main article. Section \ref{sec:related-work} are among the main successful works that inspire our idea.

\section{2D Autoencoder}
\label{sec:2D_autoencoder}

\begin{figure*}
\centering
\begin{subfigure}{.245\textwidth}
  \centering
  \includegraphics[width=\linewidth]{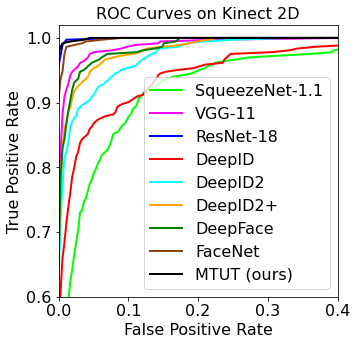}
  \label{fig:sub1}
\end{subfigure}%
\begin{subfigure}{.245\textwidth}
  \centering
  \includegraphics[width=\linewidth]{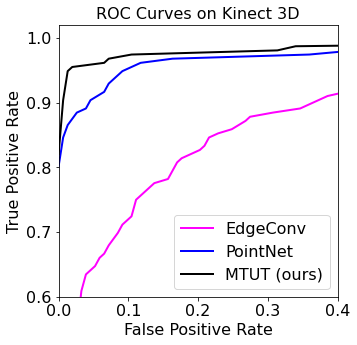}
  \label{fig:sub2}
\end{subfigure}
\begin{subfigure}{.245\textwidth}
  \centering
  \includegraphics[width=\linewidth]{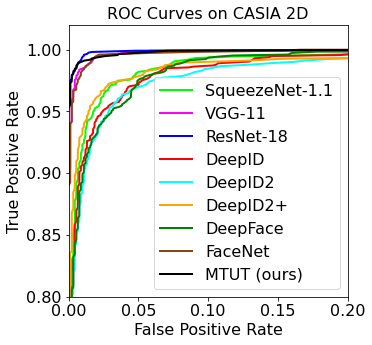}
  \label{fig:sub3}
\end{subfigure}
\begin{subfigure}{.245\textwidth}
  \centering
  \includegraphics[width=\linewidth]{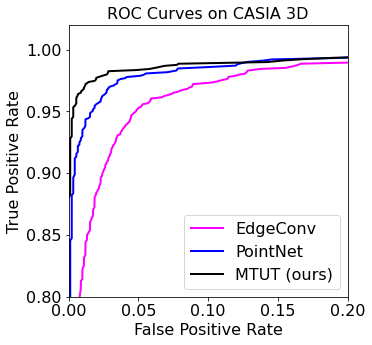}
  \label{fig:sub4}
\end{subfigure}
\begin{subfigure}{.245\textwidth}
  \centering
  \includegraphics[width=\linewidth]{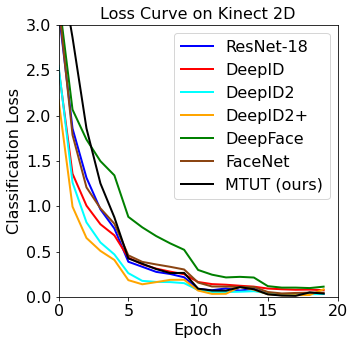}
  \label{fig:sub1}
\end{subfigure}%
\begin{subfigure}{.245\textwidth}
  \centering
  \includegraphics[width=\linewidth]{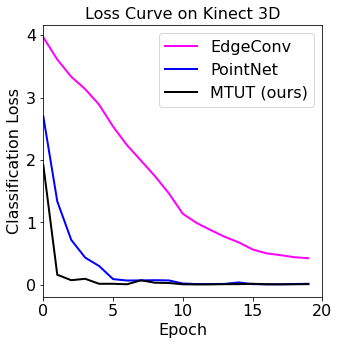}
  \label{fig:sub2}
\end{subfigure}
\begin{subfigure}{.245\textwidth}
  \centering
  \includegraphics[width=\linewidth]{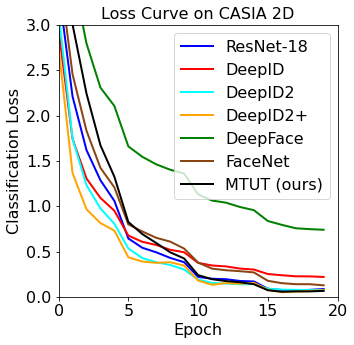}
  \label{fig:sub3}
\end{subfigure}
\begin{subfigure}{.245\textwidth}
  \centering
  \includegraphics[width=\linewidth]{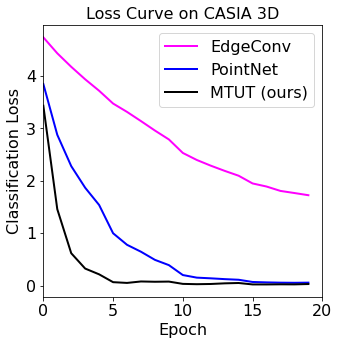}
  \label{fig:sub4}
\end{subfigure}
\caption{ROC curves (top row) and loss curves (bottom) row comparison between proposed method and baseline models on both datasets. Please note that for better visualization loss curves of VGG-11 and SqueezeNet-1.1 are not shown.}
\label{fig:roc_loss}
\end{figure*}





The upper left of Figure \ref{fig:decoder} represents the details of our 2D encoder: $enc^I$. We use ResNet-18 \cite{he2016deep} as our backbone to extract face embedding. The 2D decoder $dec^I$ is shown on the upper right of Figure \ref{fig:decoder}. $dec^I$ consists of several 2D transposed convolution layers.  Table \ref{tab:layer} shows the network configuration of our decoder network. 
We apply 6 deconvolution layers with different kernels to reconstruct face embedding into images with size of 224 $\times$ 224 $\times$ 3; a 2D batch normalization layer is also added after each deconvolution layer; ReLU activation is added afterward. The output of $dec^I$ is finally fed into a Sigmoid function to ensure that the output of 2D autoencoder is within the range of $[0,1]$ same as input image.

\begin{table}[t] 
    \centering
    \begin{tabular}{|c|c|c|c|} 
        \hline
        Layer & Kernel & Stride & Output Size \\
        \hline
        Deconv-1 & 7 $\times$ 7 & 2 &  7 $\times$ 7 $\times$ 256\\ 
        BN-1 & & & 7 $\times$ 7 $\times$ 256\\
        \hline
        Deconv-2 & 4 $\times$ 4 & 2 & 14 $\times$ 14 $\times$ 128\\ 
        BN-2 & & & 14 $\times$ 14 $\times$ 128\\
        \hline
        Deconv-3 & 4 $\times$ 4 & 2 & 28 $\times$ 28 $\times$ 128\\ 
        BN-3 & & & 28 $\times$ 28 $\times$ 128\\
        \hline
        Deconv-4 & 4 $\times$ 4 & 2 & 56 $\times$ 56 $\times$ 64\\ 
        BN-4 & & & 56 $\times$ 56 $\times$ 64\\
        \hline
        Deconv-5 & 4 $\times$ 4 & 2 & 112 $\times$ 112 $\times$ 32\\ 
        BN-5 & & & 112 $\times$ 112 $\times$ 32\\
        \hline
        Deconv-6 & 4 $\times$ 4 & 2 & 224 $\times$ 224 $\times$ 3\\ 
        Sigmoid & & & 224 $\times$ 224 $\times$ 3\\
        \hline
    \end{tabular}
    \vspace{1mm}
    \caption{The network structure of 2D decoder. Note that we also apply a ReLU layer after each convolution transpose layer section.}
    \label{tab:layer}
\end{table}

\section{3D Autoencoder} \label{sec:3D_autoencoder}
The bottom left of Figure \ref{fig:decoder} shows our 3D encoder details: $enc^P$. We employ the PointNet architecture \cite{qi2017pointnet} on 3D encoder denoted as $enc^P$. Suppose the inputs are $F$-dimensional point features with $n$ points, denoted by $\mathbf{P}=\left\{\mathbf{P}_{1}, \ldots, \mathbf{P}_{n}\right\}$, $\forall i, \mathbf{P}_i \in \mathbb{R}^{F}$. 
$F=3$ as we use $(x,y,z)$ coordinates for point features. The PointNet network regards a point cloud as a direct graph $\mathcal{G}=(\mathcal{V},\mathcal{E})$ to represent the whole point cloud structure, where $\mathcal{V}$ represents vertices of the graph, which are the points, and $\mathcal{E}$ represents the edge information obtained through $k$-nearest neighbors ($k$-NN) measured by point distances in point cloud $\mathbf{P}$.

First, the encoder learns a spatial transform to project the input point cloud to a desired space. Then, three 1D CNN layers are applied. Suppose $\mathbf{P}^l_i,i=\{1,2,\ldots,n\}$ is the input of graph convolution layer $l$, the output of layer $l$ as:

\begin{equation} \label{eq:pointnet}
    \mathbf{P}^{\prime l}_i = h_{\Theta}\left(\mathbf{P}^l_{i}\right), 
    l = 1,2,...L
\end{equation}







\noindent where $h_{\Theta}:\mathbb{R}^F  \rightarrow \mathbb{R}^{F^{\prime}}$ is a non-linear function with a set of learnable parameters $\Theta$. $F^{\prime}$ is the output dimension of graph convolution network.
The spatial transform block learns a $F \times F$ transformation matrix from features of input point cloud. The transformation of original points will project them into a space facilitating the classification task. After spatial transformation and several convolution layers, PointNet architecture extracts 1024-dimension node embedding for each vertex $v \in \mathbf{P}$. Finally, max pooling is applied to the node embeddings to obtain the final latent embedding: $\mathbf{x}^P \in \mathbb{R}^{1024}$.

Our 3D decoder $dec^P$ is illustrated on the bottom right of Figure \ref{fig:decoder}.
The last max-pooling operation for $enc^P$  is hard to reverse as the pooling is operated adaptively on all points. Therefore, instead of performing unpooling, we apply Gaussian sampling to obtain $n$ point features with a max value equal to $\mathbf{x}^P$. Suppose the input of $dec^P$ is $\hat{\mathbf{P}}^0$, then:

\begin{equation} \label{eq:gauss}
    \hat{\mathbf{P}}^0_{i,k} = min\bigg\{\mathcal{N}(0,1), \mathbf{x}^P_k\bigg\} 
\end{equation}


\noindent $\hat{\mathbf{P}}^0_{i,k}$ denotes the feature $k$ of point $i$, where $k=1,2,...,1024$ and $i=1,2,...,n$. 
The sampled point features are then fed into three 1D transposed convolutional layers along the first dimension. We set kernel size to be 1, stride to be 1 and zero-padding. 
After three deconvolution layers, the final output of $dec^P$ is a $n \times F$ matrix consisting of $n$ points.

\section{Visualization and Results} \label{sec:visual}

Figure \ref{fig:roc_loss} shows the ROC curves and classification loss curves on both datasets. As could be discovered, our proposed MTUT method has higher Area Under the Curve (AUC) value compared with other baseline methods. In addition, classification loss of our proposed method achieves global minimum earlier than majority of other baseline architectures.

\begin{figure}[htb!]
\centering
\begin{subfigure}{.45\textwidth}
  \centering
  \includegraphics[width=\linewidth]{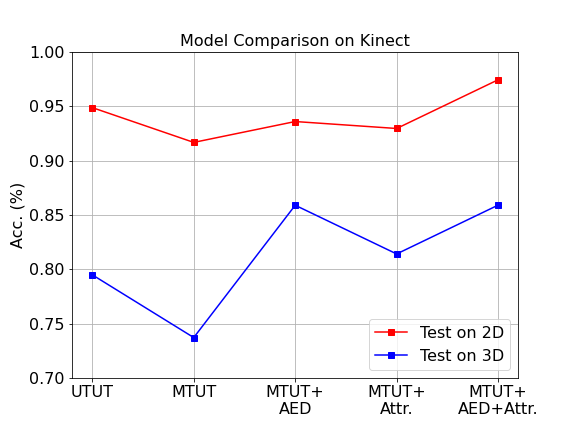}
  \label{fig:sub1}
\end{subfigure}
\hfill
\vspace{-5mm}
\begin{subfigure}{.45\textwidth}
  \centering
  \includegraphics[width=\linewidth]{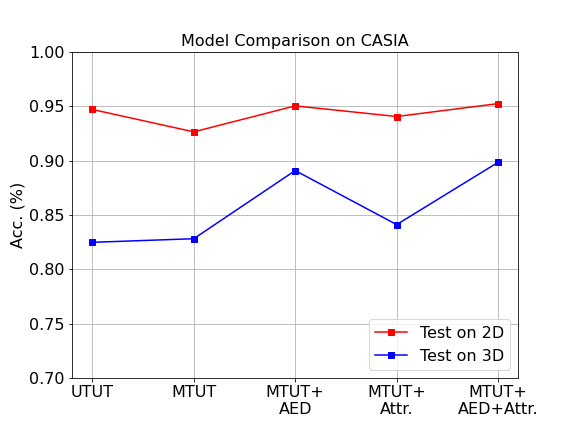}
  \label{fig:sub2}
\end{subfigure}
\vspace{-2mm}
\caption{Ablation study. Our methods are \emph{MTUT}+\emph{AED}+\emph{Attr.} UTUT stands for unimodal training unimodal testing.} 
\label{fig:ablation}
\end{figure}

\noindent\textbf{Ablation Study.} The proposed MTUT method has two additional characteristics over traditional multi-view autoencoder: (1) 
face attribute vectors (Attr.); 
(2) adaptive embedding divergence (AED) loss. 
In this section, we will evaluate the contributions of these two additional methods on face recognition accuracy. Figure \ref{fig:ablation} present our ablation study varying model components. MTUT+AED+Attr. (the rightmost) represents the full model with both (1) and (2), whereas the others represent that we \textbf{exclude} one or both of the methods.
As it could be discovered from Figure \ref{fig:ablation}, AED has significant contribution over boosting classification accuracy on both datasets. The point could be even consolidated by that the accuracy of models without AED are even less than that of backbone architecture (eg. 92.95\% vs. 94.87\% on Kinect when testing on 2D). This proves that AED successfully optimizes the encoded embeddings in order to obtain robust performance on classification network. Although adding Attr. to face embeddings would theoretically guide reconstruction of face modalities, it leads to less obvious improvement on classification accuracy compared with AED.

\section{Related Works} \label{sec:related-works}

\noindent\textbf{2D Face Recognition.} Compared with traditional face recognition methods, deep learning technique \cite{taigman2014deepface,sun2014deep,schroff2015facenet} employs larger number of layers to perform feature transformation and extraction, driving both accuracy and efficiency of face recognition up significantly. 
However, the performances are still restricted by pose and illumination variations. Compared with previous methods, our approach involves 3D data as a fundamental assistance to boost 2D model performance.


\noindent\textbf{3D Face Recognition.} 3D information of faces is crucial to face recognition modalities 
Our proposed approach is inspired by Qi \emph{et al}. \cite{qi2017pointnet}, which designs a novel neural net architecture called PointNet that is suitable for point cloud; it achieves permutation invariance by processing each point identically and independently with symmetrical function. 
Our proposed method applies the methodologies of PointNet to perform input spatial transformation and feature extraction process to obtain a high-level feature embedding. 

\noindent\textbf{Multimodel Fusion.} Multimodal fusion helps effectively integrate information from a variety of modalities with the same goal of correct classification \cite{baltruvsaitis2018multimodal}. We perform multimodal fusion because multiple modalities have access to more comprehensive information necessary to combine for more robust performance. 
However, modalities like these are well-established when resources are complete and with high quality; especially when some modality views are unseen, the feature information of the seen views cannot be effectively transferred to unseen only through modal fusion.

\noindent\textbf{Co-Learning.} Previous problems could be mitigated with the technique of co-learning, which assists the modeling of modality with poor resources by exploiting embedded information of another modality \cite{baltruvsaitis2018multimodal}. We often perform co-learning when the assisting modality is used only during training. The proposed model is primarily inspired by \cite{abavisani2019improving} that effectively achieves Multimodal Training and Unimodal Testing (MTUT) by implementing Spatiotemporal Semantic Alignment (SSA) loss function. However, \cite{abavisani2019improving} failed to fuse the characteristics of different modalities and the results are largely decided by network performance on unimodal classification. Dumpala \emph{et al}. \cite{dumpala2019audio} propose a cross-modal autoencoder (DCC-CAE) that maps the seen feature to unseen feature space through optimizing canonical correlation analysis cost function, but the power of their network would possibly be mitigated by the negative role played by any noisy modality.
{\small
\bibliographystyle{ieee}
\bibliography{egbib}
}